\documentclass[conference]{IEEEtran}
\IEEEoverridecommandlockouts

\usepackage{cite}
\usepackage{amsmath,amssymb,amsfonts}
\usepackage{algorithmic}
\usepackage{graphicx}
\usepackage{textcomp}
\usepackage{xcolor}
\usepackage{hyperref}
\usepackage{caption}       
\usepackage{subcaption}    
\usepackage{tikz}
\usepackage{comment}
\usetikzlibrary{shapes.geometric, arrows.meta, positioning, fit, backgrounds}

\def\BibTeX{{\rm B\kern-.05em{\sc i\kern-.025em b}\kern-.08em
    T\kern-.1667em\lower.7ex\hbox{E}\kern-.125emX}}

\hypersetup{
    colorlinks=true,     
    linkcolor=red,       
    citecolor=red,       
    urlcolor=red         
}
    
\begin{document}

\title{Epileptic Seizure Prediction Using Patient-Adaptive Transformer Networks\\
}

\author{\IEEEauthorblockN{Mohamed MAHDI}
\IEEEauthorblockA{\textit{ICT Engineering Student} \\
\textit{Information and Communication Engineering Department} \\
\textit{National Engineering School of Tunis}\\
\textit{Tunis El Manar University}\\
mohamed.mahdi@etudiant-enit.utm.tn
}
\and
\IEEEauthorblockN{Asma BAGHDADI}
\IEEEauthorblockA{
    \textit{REGIM-Lab, University of Sfax,}\\
    \textit{National Engineering School of Sfax (ENIS),}\\
    \textit{BP 1173, Sfax, 3038, Tunisia}\\
    \textit{National Engineering School of Tunis}\\
    \textit{Tunis El Manar University}\\
    asma.baghdadi@enit.utm.tn
}

}

\maketitle

\begin{abstract}
Epileptic seizure prediction from electroencephalographic (EEG) recordings remains challenging due to strong inter-patient variability and the complex temporal structure of neural signals. This paper presents a patient-adaptive transformer framework for short-horizon seizure forecasting. The proposed approach employs a two-stage training strategy: self-supervised pretraining is first used to learn general EEG temporal representations through autoregressive sequence modeling, followed by patient-specific fine-tuning for binary prediction of seizure onset within a 30-second horizon. To enable transformer-based sequence learning, multichannel EEG signals are processed using noise-aware preprocessing and discretized into tokenized temporal sequences. Experiments conducted on subjects from the TUH EEG dataset demonstrate that the proposed method achieves validation accuracies above 90\% and F1 scores exceeding 0.80 across evaluated patients, supporting the effectiveness of combining self-supervised representation learning with patient-specific adaptation for individualized seizure prediction.
\end{abstract}


\section{Introduction}
Epilepsy affects more than 50 million people worldwide and is characterized by recurrent, unpredictable seizures that severely impact patient safety and quality of life \cite{who2024, zhang2024review}. Because seizures occur without warning, patients face risks of injury, psychological stress, and long-term lifestyle restrictions. Current clinical systems remain largely reactive, focusing on seizure detection rather than prevention. Reliable early prediction would represent a major clinical advance, enabling timely intervention and adaptive therapies. Electroencephalogram (EEG) signals are the primary modality for seizure analysis; however, they are highly non-stationary, noisy, and strongly patient-dependent, making prediction a complex time-series modeling challenge \cite{zhang2024review}. Public resources such as the CHB-MIT and TUH EEG datasets have accelerated research while highlighting strong inter-patient variability \cite{mourad2025machine}.\\

Early approaches relied on handcrafted temporal and spectral features combined with classical machine learning classifiers\cite{usman2017epileptic}. While useful, these methods require intensive feature engineering and show limited robustness. Deep learning models, including CNNs on EEG spectrograms \cite{hussein2021epileptic} and recurrent networks such as LSTMs\cite{wu2023end}, improved automation in feature extraction but often struggle with long-range temporal dependencies. Transformer architectures address this limitation through self-attention mechanisms that efficiently model long-term relationships in sequential data\cite{zerveas2021transformer}. Generative Pre-trained Transformers (GPTs)\cite{vaswani2017attention}, have demonstrated strong performance in time-series and biomedical signal analysis due to transferable pretrained representations \cite{zerveas2021transformer}.\\

In this work, we propose a patient-specific seizure prediction framework based on a pretrained and fine-tuned GPT model trained on TUH EEG recordings. The model first learns general EEG temporal dynamics through self-supervised pretraining, then adapts to individualized pre-ictal patterns via patient-specific fine-tuning. This personalization directly addresses inter-patient variability and aligns with recent adaptive prediction strategies \cite{zhang2024scheme}. Our contributions are threefold: (1) a GPT-based architecture for EEG seizure forecasting, (2) validation of patient-specific fine-tuning on TUH EEG data, and (3) evidence that transformer attention captures meaningful pre-ictal temporal structures for early prediction.\\

The remainder of this paper is organized as follows: Section II reviews related work, Section III details the methodology, Section IV presents experiments and results, and Section V concludes the paper.\\

\noindent \textbf{The complete source code is publicly available at:} 
\href{https://github.com/Mahdyy02/gpt-eeg-transformer}{github.com/Mahdyy02/gpt-eeg-transformer}.

\section{RELATED WORK}
Recent advances in EEG analysis have been driven by deep learning and self-supervised representation learning, particularly for seizure-related tasks. It is important, however, to distinguish between \textbf{seizure events} and \textbf{epilepsy as a chronic disorder}. Seizures may result from acute causes such as fever or trauma, whereas epilepsy is defined by recurrent, unprovoked seizures supported by clinical and EEG evidence. Several studies leverage self-supervised learning to improve EEG representations. BioSerenity-E1 employs spectral tokenization and masked prediction to model temporal dependencies in EEG spectrograms \cite{bettinardi2025bioserenity}, achieving strong results in seizure detection and abnormality classification but focusing on event-level analysis rather than prediction. GMAEEG models EEG channels as graph nodes and applies masked graph autoencoding to learn spatiotemporal dynamics \cite{fu2024gmaeeg}, though its applications remain population-oriented. Masked autoencoder approaches have also been explored; Zhou and Liu reconstruct randomly masked EEG segments to learn representations useful for cognitive tasks \cite{zhou2024enhancing}. EEG2Rep further refines masking in latent space using semantic subsequence preservation, improving robustness across EEG tasks \cite{mohammadi2024eeg2rep}. However, these frameworks remain largely task-agnostic and do not explicitly address patient-specific seizure prediction. A common trend across this literature is the use of self-supervised pretraining followed by task-specific fine-tuning, underlining the importance of representation learning in EEG analysis \cite{weng2025self}. Yet most works focus on \textbf{seizure detection or classification} rather than \textbf{forecasting future seizures}. Moreover, evaluations are typically population-based, despite strong clinical variability in seizure signatures. Clinical observations confirm that EEG-based prediction is rarely used in practice, partly due to this heterogeneity. Public datasets such as CHB-MIT and TUH EEG illustrate the challenge, where seizure activity may be highly localized to specific channels, especially in focal epilepsy.\\

In contrast, our work targets \textbf{patient-specific seizure prediction}. By combining transformer architectures with self-supervised learning, we model long-range temporal dependencies and individualized pre-ictal dynamics. This approach complements existing representation-learning methods while aligning with the clinical reality of epilepsy as a personalized and recurrent condition.

\section{PROPOSED METHODOLOGY}

\subsection{Problem Definition}

Epileptic seizure prediction from electroencephalographic (EEG) signals is formulated in this work as a patient-specific, short-horizon forecasting problem. The objective is not merely to detect ongoing seizures, but to anticipate their imminent onset by identifying subtle pre-ictal dynamics embedded in multichannel brain activity.\\

\noindent Let the EEG input be defined as:

\begin{equation}
X \in \mathbb{R}^{C \times T}
\end{equation}

\noindent denote a segment of scalp EEG, where $C$ represents the number of recording channels and $T$ the temporal samples within an observation window. Each segment is extracted from continuous clinical recordings and reflects the spatial--temporal evolution of neural activity across the scalp.\\

\noindent Given an input EEG segment $X$, the model outputs a binary prediction:

\begin{equation}
y \in \{0,1\}
\end{equation}


\noindent where:

\begin{itemize}
\item $y = 1$ $\rightarrow$ A seizure will occur within the next 30 seconds
\item $y = 0$ $\rightarrow$ No seizure will occur within this horizon
\end{itemize}

\vspace{12pt}

\noindent This formulation casts seizure prediction as a supervised binary classification task operating on short temporal horizons. To construct training labels, EEG recordings are partitioned into two primary states:
\begin{itemize}
\item \textbf{Pre-ictal segments}: EEG windows preceding seizure onset within a fixed 30-second prediction horizon.
\item \textbf{Non-pre-ictal segments}: Inter-ictal or background EEG windows temporally distant from any seizure event.
\end{itemize}

\vspace{12pt}

This distinction allows the model to learn discriminative patterns that arise before seizure onset rather than features of the seizure itself. A patient-specific setting is adopted, where each model is trained and evaluated on a single subject. This reflects clinical evidence that seizure signatures, spatial propagation, and spectral characteristics vary significantly across individuals, making personalization essential for capturing subject-dependent pre-ictal biomarkers. The 30-second prediction horizon is clinically motivated, balancing early warning capability with prediction reliability. Longer horizons introduce uncertainty in pre-ictal boundaries, while very short horizons approach seizure detection instead of true forecasting. Within this framework, the system learns patient-specific mappings from multichannel EEG dynamics to imminent seizure risk, enabling actionable early warnings within clinically meaningful timeframes.

\subsection{Proposed Methodology Overview}

Figure \ref{fig:architecture} below illustrates the overall architecture of the proposed seizure prediction framework, which is designed as a multi-stage pipeline combining noise-aware preprocessing, self-supervised representation learning, and task-specific fine-tuning.\\

\begin{figure*}[t]
\centering
\resizebox{\textwidth}{!}{%
\begin{tikzpicture}[
    node distance=1cm and 0.8cm,
    >=latex,
    block/.style={rectangle, draw, fill=blue!10, text width=2.2cm, text centered, rounded corners, minimum height=0.9cm, font=\footnotesize},
    process/.style={rectangle, draw, fill=green!15, text width=2cm, text centered, rounded corners, minimum height=0.9cm, font=\footnotesize},
    decision/.style={diamond, draw, fill=blue!20, text width=1.8cm, text centered, minimum height=1cm, font=\footnotesize},
    data/.style={ellipse, draw, fill=orange!15, text width=1.8cm, text centered, minimum height=0.7cm, font=\footnotesize},
    arrow/.style={->, thick},
    label/.style={font=\tiny, text=green!60!black}
]

\node[data] (input) {EEG Signal\\$X \in \mathbb{R}^{C \times T}$};

\node[process, right=of input] (fft) {FFT\\PSD};
\node[process, right=of fft] (noise) {Noise\\Detection};
\node[process, right=of noise] (filter) {Adaptive\\Filter};
\node[block, right=of filter] (clean) {Clean\\EEG};

\node[process, below=1.2cm of clean] (norm) {Z-score\\Norm};
\node[process, left=of norm] (quant) {Quantize\\$L=512$};
\node[process, left=of quant] (token) {Tokenize\\$Z=[z_1...z_T]$};

\node[block, below=1.2cm of token, fill=green!10, text width=2.8cm, minimum height=2cm, align=left] (pretrain) {
    \textbf{\footnotesize Self-Supervised}\\
    \tiny
    • 4 layers, 4 heads\\
    • Embed: 128\\
    • Block: 512\\
    • Next-token\\
    $\mathcal{L}_{CE} + 0.1\mathcal{L}_{MSE}$
};

\node[block, right=1.5cm of pretrain, fill=blue!10, text width=2.8cm, minimum height=2cm, align=left] (finetune) {
    \textbf{\footnotesize Fine-tuning}\\
    \tiny
    • Patient-specific\\
    • 30s segments\\
    • Weighted CE\\
    • LR: $3e^{-5}$\\
    • 5000 steps
};

\node[process, right=1.5cm of finetune] (pool) {Global\\Pool};
\node[process, right=of pool] (classifier) {Classify\\Softmax};
\node[decision, right=of classifier] (output) {Predict\\$y$};
\node[block, right=of output, text width=1.8cm] (result) {$y=1$:\\Seizure\\in 30s};

\draw[arrow] (input) -- (fft);
\draw[arrow] (fft) -- (noise);
\draw[arrow] (noise) -- (filter);
\draw[arrow] (filter) -- (clean);

\draw[arrow] (clean) -- (norm);
\draw[arrow] (norm) -- (quant);
\draw[arrow] (quant) -- (token);

\draw[arrow] (token) -- (pretrain);
\draw[arrow] (pretrain) -- node[above, label] {weights} (finetune);

\draw[arrow] (finetune) -- (pool);
\draw[arrow] (pool) -- (classifier);
\draw[arrow] (classifier) -- (output);
\draw[arrow] (output) -- (result);

\node[above=0.4cm of noise, font=\small\bfseries, text=gray] {Preprocessing};
\node[above=0.4cm of quant, font=\small\bfseries, text=gray] {Tokenization};
\node[below=0.3cm of pretrain, font=\small\bfseries, text=gray] {Pre-train};
\node[below=0.3cm of finetune, font=\small\bfseries, text=gray] {Fine-tune};

\draw[dashed, gray, rounded corners] ([xshift=-0.2cm,yshift=0.3cm]fft.north west) rectangle ([xshift=0.2cm,yshift=-0.3cm]clean.south east);
\draw[dashed, gray, rounded corners] ([xshift=-0.2cm,yshift=0.3cm]token.north west) rectangle ([xshift=0.2cm,yshift=-0.3cm]norm.south east);

\end{tikzpicture}
}
\caption{Overall architecture of the proposed patient-specific seizure prediction framework. The pipeline consists of four main stages: (1) Noise-aware preprocessing using adaptive FFT-based filtering, (2) EEG tokenization with z-score normalization and quantization into 512 discrete levels, (3) Self-supervised GPT pre-training using next-token prediction with dual loss, and (4) Patient-specific supervised fine-tuning for 30-second horizon seizure prediction.}
\label{fig:architecture}
\end{figure*}
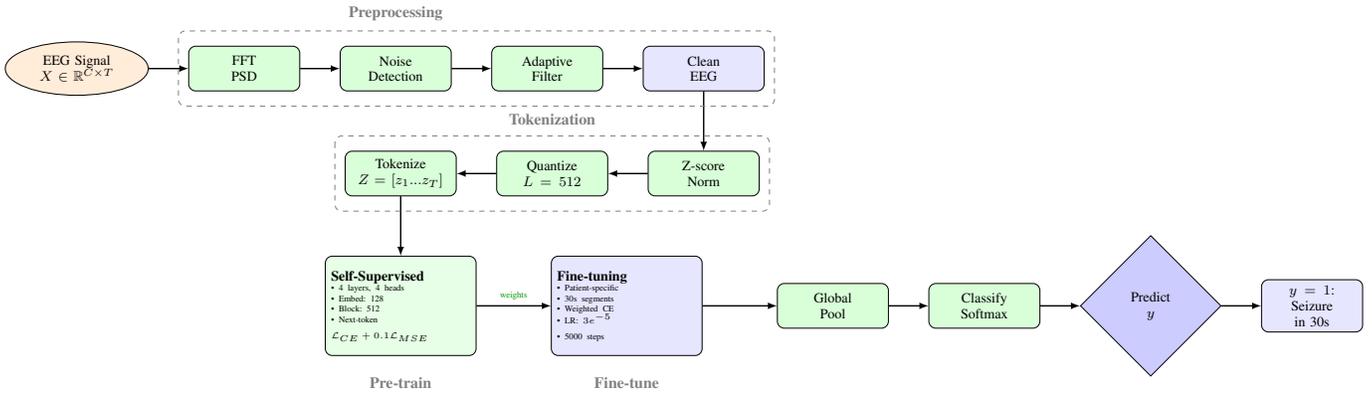

The pipeline begins with continuous multichannel scalp EEG recordings acquired from clinical monitoring sessions. These raw signals contain rich neurophysiological information but are also affected by various sources of spectral noise and acquisition artifacts. To ensure that the downstream learning stages focus on meaningful brain dynamics, the raw EEG first undergoes a dedicated preprocessing stage. In the first step, patient EEG recordings are transformed into the frequency domain using power spectral density (PSD) analysis. This transformation enables the identification of frequency components that exhibit abnormally high power combined with low variability across channels. Such spectral patterns typically correspond to uniform noise sources, including power-line interference and device-related artifacts.\\

Unlike conventional preprocessing pipelines that apply fixed notch filters at predetermined frequencies, the proposed framework performs adaptive noise detection, allowing the filtering process to be tailored to each recording session. Once noisy frequency bands are identified, adaptive notch filtering is applied selectively to remove these components while preserving physiologically relevant EEG rhythms. The output of this stage is a filtered EEG signal that retains spatial, temporal brain activity while minimizing uniform spectral contamination. Following preprocessing, the cleaned EEG signal is used to train a transformer-based foundation model in a self-supervised manner. At this stage, the model is not yet trained for seizure prediction. Instead, it learns general representations of brain dynamics directly from the filtered raw waveform. By modeling long-range temporal dependencies across EEG sequences, the transformer develops an intrinsic understanding of normal and abnormal neural patterns without relying on explicit seizure labels. After this representation learning phase, the pretrained transformer is fine-tuned for the downstream clinical task. Annotated EEG segments are introduced, where each window is labeled according to whether a seizure will occur within the predefined 30-second prediction horizon. A classification head is attached to the pretrained backbone, enabling the network to map learned EEG representations to binary seizure risk decisions. During fine-tuning, the model adapts its internal features to emphasize pre-ictal biomarkers. Because the backbone has already learned general EEG structure, the fine-tuning stage converges more efficiently.\\

The final output of the pipeline is a binary prediction indicating imminent seizure risk. This prediction is generated for each EEG segment and can be used to trigger clinical alarms or intervention systems within actionable timeframes.\\

Overall, the methodology integrates adaptive signal processing with large-scale sequence modeling. By coupling noise-aware preprocessing with transformer-based representation learning, the framework aims to maximize sensitivity to pre-ictal dynamics while maintaining robustness to real-world EEG noise conditions.

\subsection{Noise-Aware EEG Preprocessing}

Clinical EEG recordings are often contaminated by noise that can mask subtle pre-ictal dynamics. Common artifacts include broadband spectral noise, power-line interference \textbf{(50/60 Hz)}, and device-related disturbances from electrodes or amplifiers. These sources typically appear as high-power, low-variance frequency components shared across channels and can degrade seizure prediction performance if not addressed. To mitigate this effect, the preprocessing pipeline incorporates automatic noise detection. Power spectral density is used to estimate frequency energy, while cross-channel variability helps distinguish neural activity from uniform noise. Components showing both high power and low variance are flagged for filtering. Instead of applying fixed notch filters, the framework employs \textbf{adaptive notch filtering}, targeting only the identified contaminated frequencies. This selective approach preserves physiologically meaningful low-frequency activity critical for pre-ictal pattern recognition. By suppressing uniform spectral artifacts while retaining relevant neural dynamics, the preprocessing stage improves EEG signal quality and supports more reliable short-horizon seizure forecasting.

\subsection{Tokenization and EEG Sequence Representation}

To enable transformer-based learning on electroencephalographic signals, the continuous EEG waveform is reformulated as a discrete temporal token sequence. Instead of relying on handcrafted spectral or statistical features, the proposed approach models raw neural activity directly at the sample level, allowing the architecture to learn representations end-to-end from the signal itself. Let $X$ our EEG segment, for sequence modeling, the multichannel recordings are first flattened into a one-dimensional temporal stream by concatenating all channels for each time step in order. Each individual sample is then treated as an atomic observation unit to be converted into a token. Because transformers operate on discrete vocabularies, amplitude values must be normalized and quantized prior to tokenization. The raw signal is first standardized using z-score normalization:

\begin{equation}
X_{\text{norm}} = \frac{X - \mu}{\sigma}
\end{equation}

\noindent where $\mu$ and $\sigma$ denote the global mean and standard deviation of the EEG recording. This step removes amplitude scaling differences across sessions and stabilizes training. To limit the influence of extreme outliers, the normalized signal is clipped within a bounded interval:

\begin{equation}
X_{\text{clip}} = \text{clip}(X_{\text{norm}}, -k\sigma, +k\sigma)
\end{equation}

\noindent where $k$ is a clipping factor (set to 5 in this work). The clipped values are then linearly mapped into the unit interval:

\begin{equation}
X_{01} = \frac{X_{\text{clip}} + k\sigma}{2k\sigma}
\end{equation}

This bounded representation is subsequently discretized into $L$ quantization levels:

\begin{equation}
\text{Token} = \lfloor X_{01} \times (L - 1) \rfloor
\end{equation}

\noindent where $L$ denotes the vocabulary size (i.e., the number of discrete amplitude bins). Each EEG sample is therefore converted into a single integer token in the range $[0, L-1]$. This design establishes a \textbf{one-token-per-sample representation}, enabling the EEG waveform to be processed analogously to a language sequence. Long temporal dependencies can thus be captured through transformer self-attention without requiring manual feature engineering. Unlike conventional EEG pipelines that rely on spectrograms, wavelets, or band-power statistics, the proposed tokenization preserves the raw temporal resolution of the signal. This allows the model to learn both micro-scale waveform morphology and macro-scale temporal structure within large context windows. Positional embeddings are added to the token embeddings to encode temporal ordering, ensuring that the model distinguishes early from late samples within each observation window. Consequently, the transformer learns joint amplitude--temporal representations directly from quantized neural dynamics. To further constrain reconstruction fidelity, training is guided by a dual-objective loss combining categorical prediction and waveform regression. In addition to cross-entropy over token classes, a mean-squared error term is computed after dequantizing predicted token probabilities back into waveform space. This auxiliary supervision encourages the model to preserve physiologically meaningful signal morphology rather than only discrete token accuracy.

\subsubsection*{Hyperparameter Summary (Tokenization \& Sequence Modeling)}

Table~\ref{tab:tokenization_hyperparams} summarizes the key hyperparameters used for tokenization and sequence modeling. The quantization levels determine the number of discrete amplitude bins, where higher values preserve waveform precision but increase model complexity. The block size sets the context window length processed by the transformer, controlling how much temporal history the model can attend to. Z-score normalization stabilizes gradients and training convergence, while the clipping factor limits extreme values to prevent outliers from dominating quantization. Each EEG sample is tokenized individually, preserving full temporal resolution, and embedded into vectors of dimension 128. Learned positional embeddings encode temporal order for sequence awareness. Gradient accumulation simulates a larger batch size to improve training stability under memory constraints. Finally, the dual loss weight balances waveform reconstruction and encourages physiologically realistic outputs.

\begin{table}[htbp]
\centering
\caption{Hyperparameters for Tokenization and Sequence Modeling}
\label{tab:tokenization_hyperparams}
\begin{tabular}{ll}
\hline
\textbf{Hyperparameter} & \textbf{Value} \\
\hline
Quantization Levels (L) & 512 \\
Block Size & 512 samples \\
Normalization & Z-score \\
Clipping Factor (k) & 5 \\
Token per Sample & 1 \\
Embedding Dimension & 128 \\
Positional Embeddings & Learned \\
Gradient Accumulation & 8 steps \\
Dual Loss Weight (MSE) & 0.1 \\
\hline
\end{tabular}
\end{table}

\subsection{Self-Supervised Transformer Pretraining}

To learn intrinsic electroencephalographic dynamics prior to seizure prediction, a self-supervised pretraining stage is introduced. This phase aims to model the temporal structure of EEG signals without relying on seizure annotations, allowing the architecture to acquire generalizable neural representations directly from raw brain activity.\\

Following tokenization, the EEG waveform is represented as a discrete sequence of amplitude tokens:

\begin{equation}
Z = [z_1, z_2, \ldots, z_T]
\end{equation}

\noindent where each token corresponds to a quantized EEG sample. The transformer is trained using an autoregressive next-sample prediction objective. Given a context window of length $B$, the model receives:

\begin{equation}
[z_t, z_{t+1}, \ldots, z_{t+B-1}]
\end{equation}

\noindent and learns to predict the subsequent token:

\begin{equation}
z_{t+B}
\end{equation}

These transitions encode neural rhythms, waveform morphology, and temporal dependencies spanning multiple physiological time scales.\\

The core optimization objective is based on categorical cross-entropy loss applied to the predicted token distribution. Let $\hat{p}_t$ denote the predicted probability vector over the vocabulary at time step $t$, and $z_t$ the true token. The primary loss is defined as:

\begin{equation}
\mathcal{L}_{\text{CE}} = -\sum_t \log \hat{p}_t(z_t)
\end{equation}

Minimizing this objective encourages the transformer to accurately model discrete amplitude transitions and learn the generative structure of EEG waveforms.\\

However, discrete token prediction alone may not guarantee faithful reconstruction of the underlying continuous signal. To preserve waveform fidelity, an auxiliary regression objective is introduced. After computing the softmax probabilities, the expected quantized amplitude level is estimated as:

\begin{equation}
\hat{e}_t = \sum_k \hat{p}_t(k) \cdot k
\end{equation}

\noindent where $k$ indexes quantization levels. This expected level is then dequantized back into waveform space using the inverse normalization mapping:

\begin{equation}
\hat{x}_t = \text{Dequantize}(\hat{e}_t)
\end{equation}

The reconstructed waveform is compared with the ground-truth signal $x_t$ using mean squared error:

\begin{equation}
\mathcal{L}_{\text{MSE}} = \frac{1}{T} \sum_t (\hat{x}_t - x_t)^2
\end{equation}

The final training objective combines discrete and continuous supervision:

\begin{equation}
\mathcal{L}_{\text{total}} = \mathcal{L}_{\text{CE}} + \lambda \cdot \mathcal{L}_{\text{MSE}}
\end{equation}

\noindent where $\lambda$ is a weighting coefficient controlling the contribution of waveform reconstruction. In our implementation, $\lambda = 0.1$.\\

This dual-loss formulation enables the model to learn both symbolic token transitions and physiologically meaningful waveform morphology. The pretraining transformer consists of stacked self-attention blocks with learned token and positional embeddings, allowing latent representations to preserve temporal ordering. Through multi-head self-attention, the model captures long-range EEG dependencies, including oscillatory rhythms, transient events, and pre-ictal trends. To stabilize training on long recordings, signals are processed using fixed-length context windows \textbf{(block size = 512)}, with gradient accumulation and mixed-precision training applied to address GPU memory constraints and accelerate optimization. Pretraining is conducted in a patient-specific setting, enabling the transformer to learn individualized neural dynamics and develop rich representations that are later specialized for seizure prediction during supervised fine-tuning.
\subsection{Fine-Tuning Strategy}

Following self-supervised pretraining, the transformer backbone is adapted to the clinical task of imminent seizure forecasting through supervised fine-tuning. The objective of this stage is to map learned neural representations to short-horizon seizure risk, enabling actionable early warning in real clinical settings.\\

Given a multichannel EEG segment, the model predicts whether a seizure will occur within the immediate future. Let:

\begin{equation}
X \in \mathbb{R}^{C \times T}
\end{equation}

\noindent denote a 30-second EEG window, where $C$ represents the number of scalp channels and $T$ the number of temporal samples. After preprocessing and normalization, this segment is forwarded through the pretrained transformer encoder to obtain a latent representation:

\begin{equation}
H = \text{Transformer}(X)
\end{equation}

\noindent where $H \in \mathbb{R}^{T \times d}$ captures temporal dependencies and cross-channel interactions learned during pretraining.\\

To perform seizure prediction, a classification head is appended to the backbone. Temporal features are first aggregated using global average pooling:

\begin{equation}
\bar{h} = \frac{1}{T} \sum_t H_t
\end{equation}

This pooled embedding summarizes the global neural state of the observation window. It is subsequently passed through a feed-forward classification module producing logits:

\begin{equation}
\hat{y} = \text{Classifier}(\bar{h})
\end{equation}

The final output is a probability score:

\begin{equation}
p = \text{Softmax}(\hat{y})
\end{equation}

\noindent where $p \in [0,1]^2$ encodes the likelihood of imminent seizure occurrence. A positive label $(y = 1)$ indicates that a seizure begins within the next 30 seconds, whereas a negative label $(y = 0)$ denotes the absence of seizure activity within that prediction horizon.\\

\subsubsection{Temporal Windowing and Label Construction}

Training samples are generated using a sliding-window segmentation strategy applied to continuous EEG recordings. Each 30-second segment is extracted with 50\% overlap to increase dataset density and capture transitional neural dynamics.\\

Let $t_{\text{end}}$ denote the end time of a segment. The associated prediction window is defined as:

\begin{equation}
[t_{\text{end}}, t_{\text{end}} + 30\text{ s}]
\end{equation}

If any clinically annotated seizure onset falls within this interval, the segment is labeled positive. Otherwise, it is labeled negative. This labeling strategy ensures that the model learns pre-ictal signatures rather than ictal seizure morphology. Clinical annotations are parsed from expert-reviewed EEG reports, and seizure intervals are merged across channels to form unified onset windows. This prevents redundant labeling when seizures propagate spatially across electrodes.\\

\subsubsection{Transfer Learning and Backbone Reuse}

Fine-tuning leverages the pretrained transformer weights obtained during the self-supervised stage. Token embeddings and positional encodings are transferred directly, while the encoder layers retain learned temporal attention patterns. New components are introduced to adapt the model to supervised prediction. First, a linear projection maps multichannel EEG inputs into the embedding dimension. Second, a classification head composed of LayerNorm, GELU activation, dropout regularization, and fully connected layers produces binary logits. Depending on the experimental configuration, the transformer backbone may be either frozen (feature extraction mode) or fully trainable (end-to-end fine-tuning). In this work, full fine-tuning is adopted to allow seizure-specific adaptation of attention weights. Optimization is performed using weighted cross-entropy loss to address class imbalance between pre-ictal and non-pre-ictal segments:

\begin{equation}
\mathcal{L} = -w_1 y \log(p) - w_0 (1 - y) \log(1 - p)
\end{equation}

where $w_1$ and $w_0$ are inverse-frequency class weights computed from the training distribution.\\

\subsubsection{Patient-Specific Training Paradigm}

A strictly patient-specific training strategy is employed throughout this study. Models are trained and evaluated independently for each subject using only their own EEG recordings. No cross-patient data mixing or transfer is performed during supervised learning. This design prevents information leakage and ensures that reported performance reflects realistic clinical deployment conditions. The rationale for patient-specific modeling is grounded in neurophysiological variability. Seizure onset patterns differ substantially across individuals in terms of spectral content, spatial propagation pathways, and pre-ictal duration. Population-level models often fail to capture these idiosyncratic biomarkers, leading to degraded prediction reliability. By contrast, patient-specific transformers learn subject-dependent neural signatures, enabling more precise anticipation of seizure transitions. This paradigm also aligns with clinical practice, where prediction systems are calibrated per patient following individualized monitoring sessions.\\

\subsubsection{Fine-Tuning Hyperparameters}

Table~\ref{tab:finetuning_hyperparams} lists the fine-tuning hyperparameters used for seizure prediction. Each EEG segment spans \textbf{30 seconds}, with a prediction horizon of the same length, defining the input context and forecast window. EEG signals are resampled at \textbf{250 Hz}, producing sequence lengths of \textbf{7,500 samples} for transformer input. The batch size is set to 16, with gradient accumulation over 4 steps to optimize memory usage and maintain training stability. A learning rate of $3 \times 10^{-5}$ is used to prevent catastrophic forgetting during fine-tuning. The transformer encoder comprises 4 layers with 4 attention heads and embeddings of dimension 128, enabling rich temporal representation. Dropout of 0.2 helps reduce overfitting, while the weighted cross-entropy loss addresses class imbalance and improves sensitivity. Backbone freezing can be optionally applied to control feature adaptation, and mixed-precision (FP16) training is enabled for computational efficiency.

\begin{table}[htbp]
\centering
\caption{Fine-Tuning Hyperparameters}
\label{tab:finetuning_hyperparams}
\begin{tabular}{ll}
\hline
\textbf{Hyperparameter} & \textbf{Value} \\
\hline
Segment Duration & 30 s \\
Prediction Horizon & 30 s \\
Sampling Rate & 250 Hz \\
Sequence Length & 7,500 samples \\
Batch Size & 16 \\
Gradient Accumulation & 4 steps \\
Learning Rate & $3 \times 10^{-5}$ \\
Transformer Layers & 4 \\
Attention Heads & 4 \\
Embedding Dimension & 128 \\
Dropout & 0.2 \\
Loss Function & Weighted Cross-Entropy \\
Backbone Freezing & Optional \\
Mixed Precision & Enabled \\
\hline
\end{tabular}
\end{table}

\section{EXPERIMENTAL RESULTS}

This section presents dataset statistics, training behavior, and seizure prediction performance in a patient-specific setting, where each model is trained and evaluated on a single subject. These experiments are designed as a \textbf{proof-of-concept} rather than a large population study. The goal is to show that the proposed framework, combining self-supervised pretraining and patient-specific fine-tuning, can work effectively on EEG data. Self-supervised pretraining was conducted for \textbf{5,000 epochs}, followed by supervised fine-tuning for \textbf{5,000 optimization steps}. We report per-patient results to reflect the strong variability between individuals in EEG signals. We present dataset characteristics, training curves, and prediction results, along with representative alarm timelines.
\subsection{Dataset characteristics}

Table~\ref{tab:dataset_aaaaaaac}, Table~\ref{tab:dataset_aaaaabnn}, and Table~\ref{tab:dataset_aaaaadpj} summarize the recordings used in our experiments. The tables report the number of sessions per patient, total recording hours, annotated seizure counts, number of 30-second segments generated (50\% overlap for training split generation or 75\% overlap for inference), fraction of segments labeled pre-ictal, the sampling frequency used after resampling, and the standardized channel count for each patient. The table shows heterogeneity in data volume and class balance across patients. Some folders contained only a few sessions and few seizures, while others (e.g., \texttt{s005\_2006} and \texttt{s002\_2004}) include more annotated events and a larger proportion of pre-ictal segments. This variability motivates the \textbf{patient-specific training strategy} described in Section III-A.

\begin{table*}[htbp]
\centering
\caption{Dataset summary for patient \texttt{aaaaaaac}}
\label{tab:dataset_aaaaaaac}
\begin{tabular}{lcccccccc}
\hline
\textbf{Patient aaaaaaac} & \textbf{Sessions} & \textbf{Recording Hours} & \textbf{Seizures} & \textbf{Segments} & \textbf{Pre-ictal \%} & \textbf{Sampling Freq (Hz)} & \textbf{Channels} \\
\hline
s001\_2002 & 2 & 0.15 & 2 & 29 & 89.66 & 250 & 33 \\
s002\_2002 & 1 & 0.07 & 1 & 14 & 92.86 & 250 & 33 \\
s004\_2002 & 2 & 0.26 & 0 & 56 & 0.0 & 250 & 33 \\
s005\_2002 & 4 & 0.35 & 0 & 71 & 0.0 & 250 & 33 \\
\hline
\end{tabular}
\end{table*}

\begin{table*}[htbp]
\centering
\caption{Dataset summary for patient \texttt{aaaaabnn}}
\label{tab:dataset_aaaaabnn}
\begin{tabular}{lcccccccc}
\hline
\textbf{patient aaaaabnn} & \textbf{Sessions} & \textbf{Recording Hours} & \textbf{Seizures} & \textbf{Segments} & \textbf{Pre-ictal \%} & \textbf{Sampling Freq (Hz)} & \textbf{Channels} \\
\hline
s001\_2004 & 1 & 0.33 & 0 & 75 & 0.0 & 250 & 33 \\
s002\_2004 & 6 & 0.35 & 4 & 68 & 54.41 & 250 & 128 \\
s003\_2004 & 2 & 0.25 & 0 & 53 & 0.0 & 250 & 32 \\
s004\_2004 & 4 & 0.35 & 0 & 70 & 0.0 & 250 & 32 \\
\hline
\end{tabular}
\end{table*}

\begin{table*}[htbp]
\centering
\caption{Dataset summary for patient \texttt{aaaaadpj}}
\label{tab:dataset_aaaaadpj}
\begin{tabular}{lcccccccc}
\hline
\textbf{Patient aaaaadpj} & \textbf{Sessions} & \textbf{Recording Hours} & \textbf{Seizures} & \textbf{Segments} & \textbf{Pre-ictal \%} & \textbf{Sampling Freq (Hz)} & \textbf{Channels} \\
\hline
s003\_2006 & 1 & 0.37 & 0 & 84 & 0.0 & 250 & 41 \\
s005\_2006 & 6 & 0.42 & 7 & 81 & 27.16 & 250 & 32 \\
\hline
\end{tabular}
\end{table*}

\subsection{Training behaviour and convergence}

\begin{figure}[htbp]
\centering

\begin{subfigure}{0.45\textwidth}
\centering
\includegraphics[width=\linewidth]{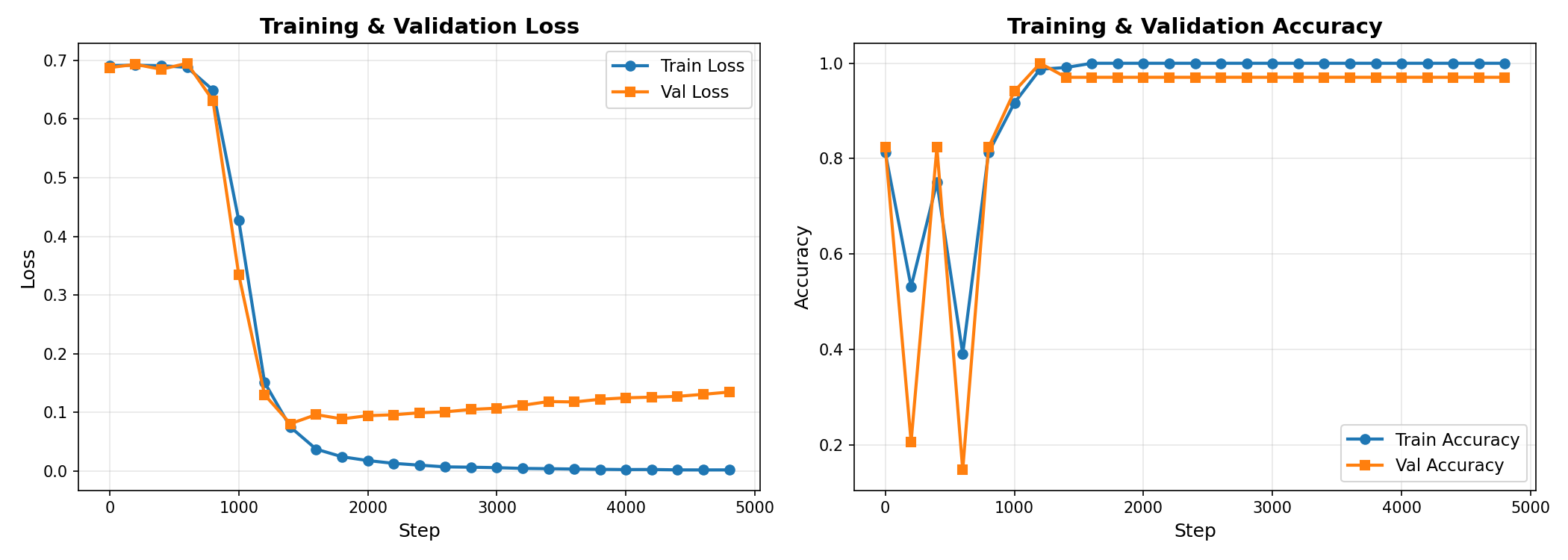}
\caption{Patient \texttt{aaaaaaac}}
\end{subfigure}

\vspace{0.1cm}

\begin{subfigure}{0.45\textwidth}
\centering
\includegraphics[width=\linewidth]{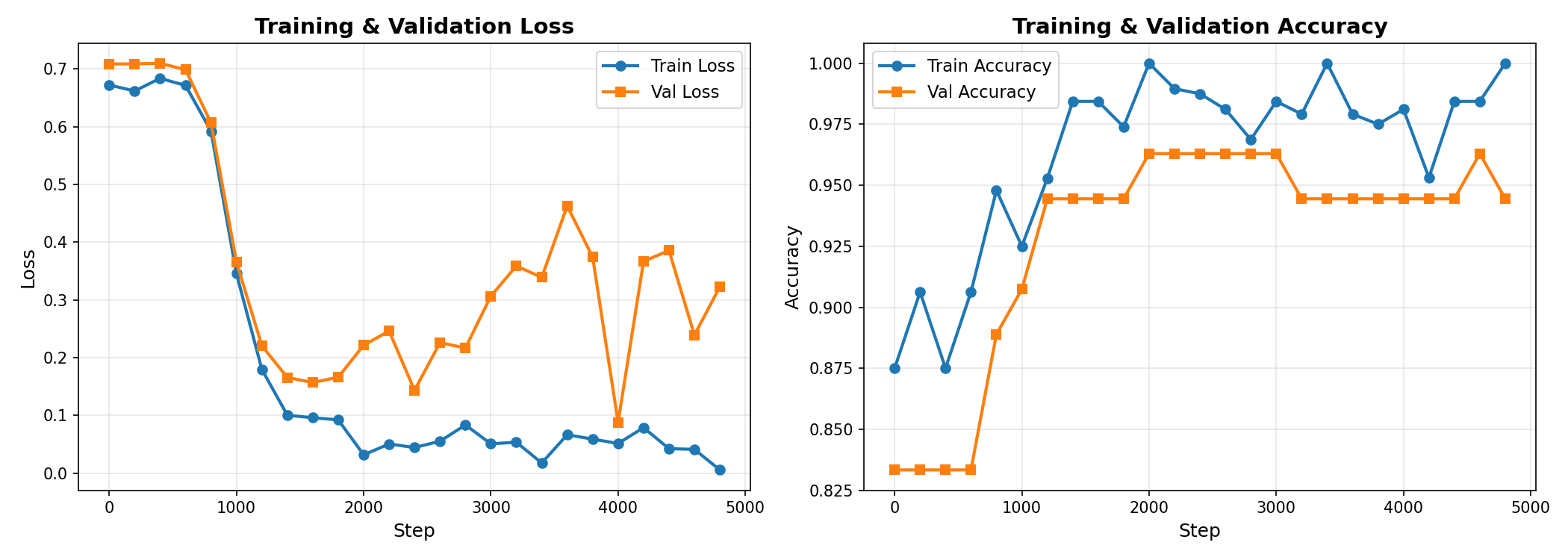}
\caption{Patient \texttt{aaaaabnn}}
\end{subfigure}

\vspace{0.1cm}

\begin{subfigure}{0.45\textwidth}
\centering
\includegraphics[width=\linewidth]{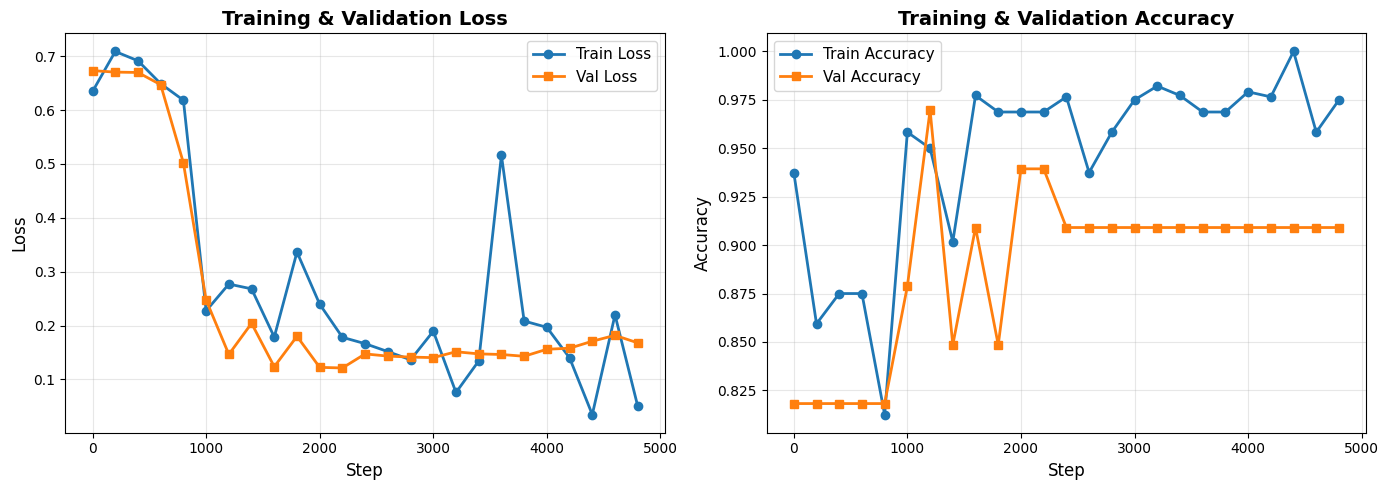}
\caption{Patient \texttt{aaaaadpj}}
\end{subfigure}

\caption{Training convergence — Training and validation loss/accuracy curves for three patient-specific fine-tuning runs. Both pretraining and fine-tuning were run for 5,000 steps; gradient accumulation and mixed precision were used to maximize effective batch size.}
\label{fig:training_convergence}
\end{figure}

All pretraining and fine-tuning runs used the transformer backbone described in Section III-F. Figure~\ref{fig:training_convergence} displays the training and validation curves (loss and accuracy) for the three main experimental runs. Curves show rapid decrease of training loss and steady validation performance after a few hundred to a few thousand steps.

\begin{table*}[htbp]
\centering
\caption{Patient-specific seizure prediction performance}
\label{tab:training_validation}
\begin{tabular}{lcccccccc}
\hline
\textbf{Patient} & \textbf{Train Acc} & \textbf{Val Acc} & \textbf{Precision} & \textbf{Recall} & \textbf{F1} & \textbf{FAR (h\textsuperscript{-1})} & \textbf{Pdelay (s)} & \textbf{Sens. (\%)} \\
\hline
aaaaaaac & 1.0000 & 0.9706 & 1.0000 & 0.8333 & 0.9091 & 0.00 & 6.89 & 100.0 \\
aaaaabnn & 1.0000 & 0.9444 & 0.8750 & 0.7778 & 0.8235 & 0.00 & 12.55 & 100.0 \\
aaaaadpj & 1.0000 & 0.9394 & 0.7500 & 1.0000 & 0.8571 & 13.82 & 30.00 & 100.0 \\
\hline
\end{tabular}
\end{table*}

\subsection{Seizure-prediction performance and alarm behaviour}

\begin{figure}[htbp]
\centering

\begin{subfigure}{0.4\textwidth}
\centering
\includegraphics[width=\linewidth]{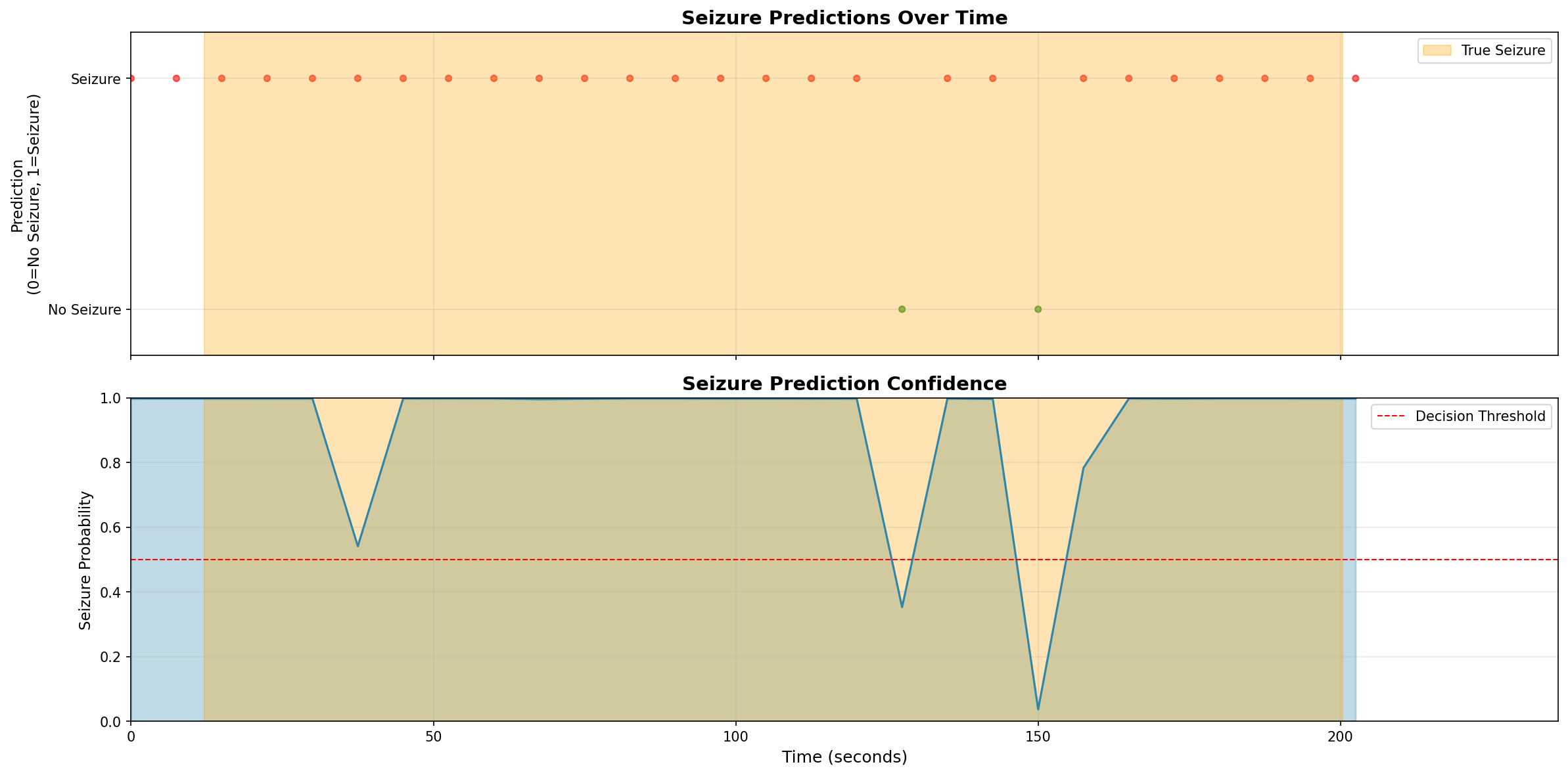}
\caption{Patient \texttt{aaaaaaac}}
\end{subfigure}

\vspace{0.1cm}

\begin{subfigure}{0.4\textwidth}
\centering
\includegraphics[width=\linewidth]{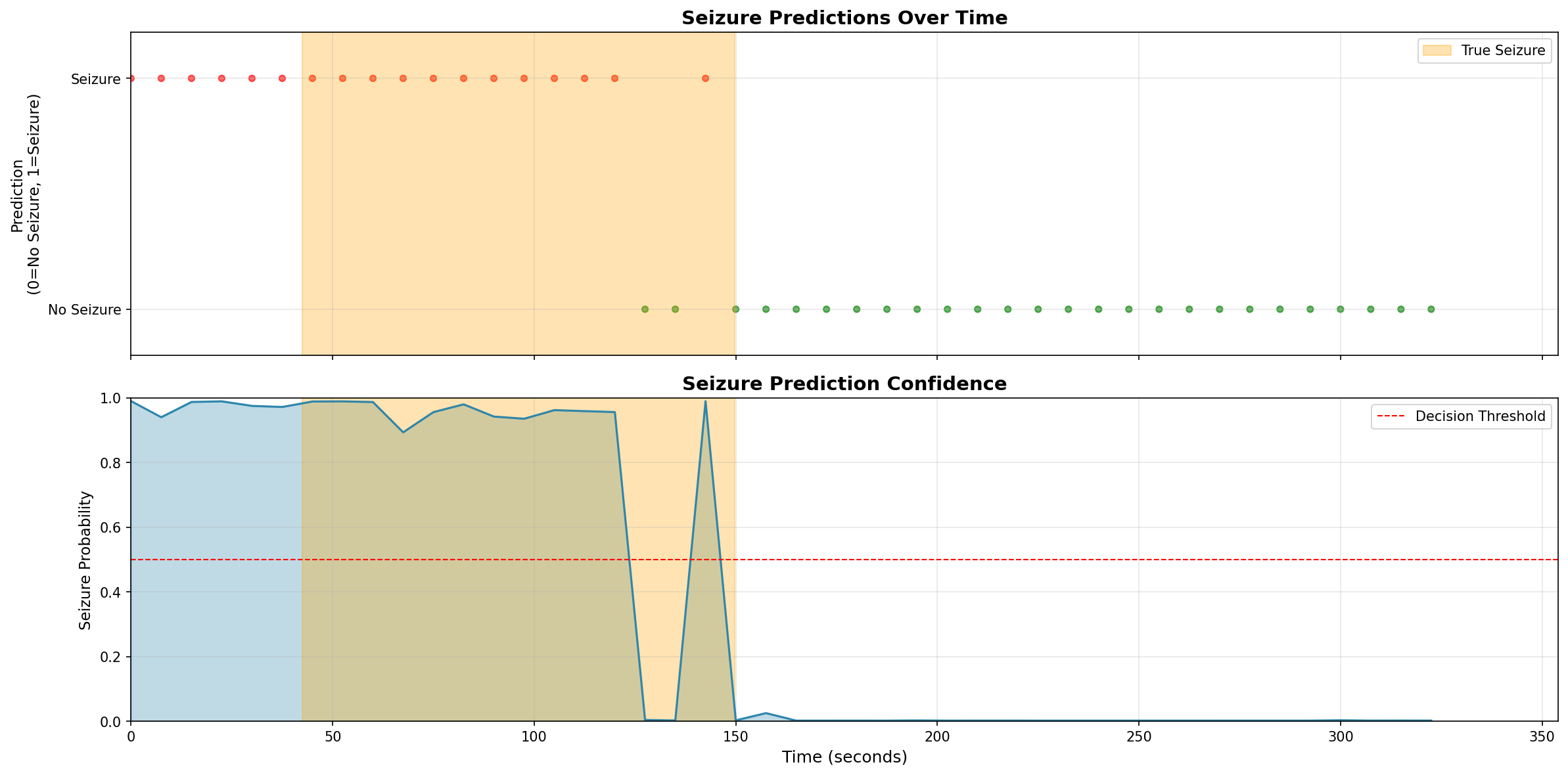}
\caption{Patient \texttt{aaaaabnn}}
\end{subfigure}

\vspace{0.1cm}

\begin{subfigure}{0.4\textwidth}
\centering
\includegraphics[width=\linewidth]{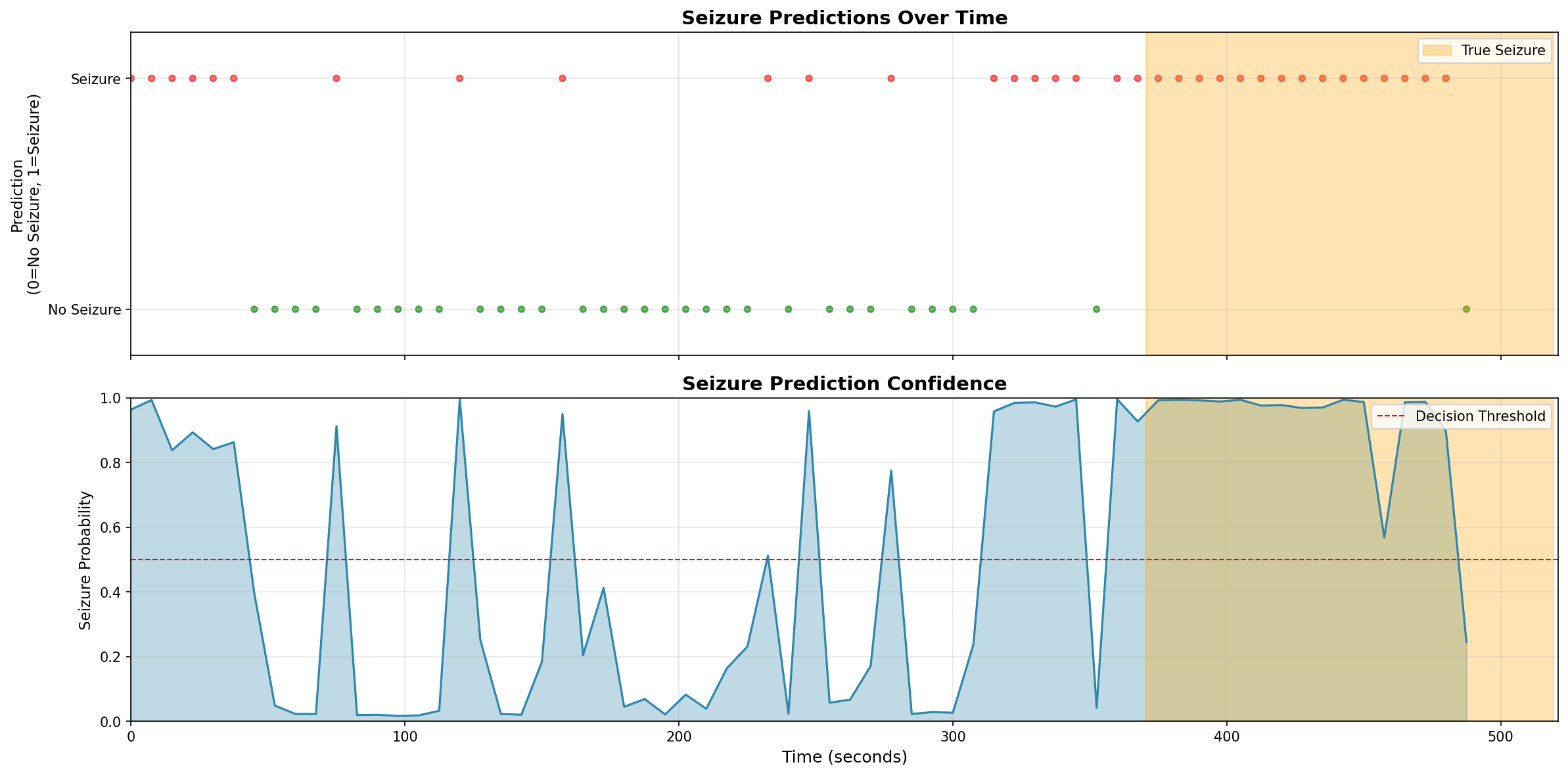}
\caption{Patient \texttt{aaaaadpj}}
\end{subfigure}

\caption{Representative alarm timeline and confidence curve -- Top: binary alarms (red) vs non-alarm (green) over time with ground-truth seizure windows highlighted in orange. Bottom: model seizure probability (confidence) over the same recording. Vertical axis shows probability in [0,1]; dashed line indicates decision threshold.}
\label{fig:alarm_timeline}

\end{figure}

Continuous recordings were segmented into overlapping 30-second windows (75\% overlap during inference), producing a time series of seizure probabilities and binary alarms (threshold = 0.5). Representative prediction timelines are shown in Figure~\ref{fig:alarm_timeline}.

\subsection{Discussion}

The experimental results indicate that our proposed framework can effectively capture short-term predictive patterns in EEG signals, even with substantial inter-patient variability and limited annotated pre-ictal segments. Across the three patients, the model achieved high validation accuracies \textbf{(around 0.94–0.97)} and robust F1 scores \textbf{(0.82–0.91)}, while detecting all seizure events. False alarm rates varied across subjects, remaining very low for two patients and higher for the third. This is partly due to the short prediction horizon and the overlapping evaluation windows, which increase temporal resolution but can produce multiple alarms for the same preictal transition. Prediction delays ranged from a few seconds up to 30 seconds, reflecting the constraints of short recording durations. It should be emphasized that this study was designed as a proof-of-concept demonstration rather than a benchmark comparison. No baseline models were included; the focus is on evaluating the feasibility of a self-supervised pretrained transformer that can adapt to individual patient dynamics and provide a flexible foundation for EEG-based predictions. Despite the small cohort and limited data, these results suggest that transformer-based models are capable of learning complex temporal dependencies in EEG signals and delivering clinically meaningful short-horizon forecasts. The observations motivate further studies on larger datasets with longer pre-ictal recordings, as well as strategies such as threshold tuning and temporal aggregation to reduce false alarms without compromising early detection performance.

\section{CONCLUSION}

Epileptic seizures remain highly unpredictable, posing a major burden for patients. This work shows that seizure onset can be anticipated by modeling EEG signals as structured temporal sequences rather than noisy recordings. Using transformer-based architectures, the proposed patient-specific framework combines self-supervised pretraining with individualized fine-tuning, enabling the model to learn general brain dynamics before specializing in patient-dependent pre-ictal patterns. This strategy achieved validation accuracies above \textbf{90\%} and F1 scores exceeding \textbf{0.80} across subjects. Beyond quantitative performance, results indicate that transformers effectively capture long-range temporal dependencies preceding seizures, generating probability trajectories and alarms within a 30-second prediction horizon. Such capability supports a shift from reactive seizure detection to proactive risk mitigation. Future work should focus on large-scale validation, optimizing the sensitivity–false alarm trade-off, and integrating wearable EEG systems for real-time monitoring.

\bibliographystyle{IEEEtran}
\bibliography{references}

\end{document}